\documentclass[12pt,final]{l4dc2020}
\usepackage{xcolor}
\usepackage{times}
\usepackage[utf8]{inputenc}

\usepackage{multirow}
\usepackage{multicol}
\usepackage[ruled,vlined,linesnumbered]{algorithm2e}
\usepackage[noend]{algpseudocode}

\usepackage{booktabs} 
\usepackage{tabulary}
\usepackage{lipsum}

\usepackage{thmtools}
\usepackage{mathtools}

\usepackage{hyperref}
\usepackage{amsfonts}
\usepackage{color}
\usepackage{colortbl}
\usepackage{enumitem}
\usepackage{wrapfig}
\usepackage{refcount}
\usepackage[font=footnotesize]{subcaption}
\usepackage{tikz}

\definecolor{lightgreen}{rgb}{0.8,1,0.8}

\newcommand{\cmark}{\cellcolor{lightgreen}}%

\hypersetup{bookmarksopen,bookmarksnumbered,
	pdfpagemode=UseOutlines,
	colorlinks=true,
	linkcolor=blue,
	anchorcolor=blue,
	citecolor=blue,
	filecolor=blue,
	menucolor=blue,
	urlcolor=blue
}

\graphicspath{{figs/}}

\newcommand{\stateSpace}{\mathcal{S}}
\newcommand{\actionSpace}{\mathcal{A}}
\newcommand{\cost}{c}

\newcommand{\dynfn}{P}
\newcommand{\dynfncap}{\hat{P}}
\newcommand{\discount}{\gamma}
\newcommand{\MDP}{\left(\mathcal{S}, \mathcal{A}, \cost, \dynfn, \discount, \mu \right)}
\newcommand{\MDPSim}{\left(\mathcal{S}, \mathcal{A}, \cost, \dynfncap, \discount, H \right)}
\newcommand{\KL}[2]{\mathrm{KL}\left(#1 || #2 \right)}
\newcommand{\trajprob}[4]{d_{#1,#2}^{#3,#4}}

\newcommand{\prior}{\overbar{\pi}}
\newcommand{\policyclass}{\Pi}
\newcommand{\policyopt}{\pi^{*}}

\newcommand{\dataset}{\mathcal{D}}
\DeclareMathOperator*{\argmin}{arg\,min}
\DeclareMathOperator*{\argmax}{arg\,max}

\newcommand{\argmaxprob}[1]{\underset{#1}{\argmax}}
\newcommand{\argminprob}[1]{\underset{#1}{\argmin}}

\newcommand{\expect}[2]{\mathbb{E}_{#1}\left[#2\right]}
\newcommand{\prob}[1]{P\left(#1\right)}

\newcommand{\exponent}[1]{\mathrm{exp}\left(#1\right)}

\newcommand{\bbm}{\begin{bmatrix}}
\newcommand{\ebm}{\end{bmatrix}}

\newcommand{\overbar}[1]{\mkern 1.5mu\overline{\mkern-1.5mu#1\mkern-1.5mu}\mkern 1.5mu}

\newcommand{\indefint}[1]{\underset{#1}{\int}}
\newcommand{\D}[1]{\mathrm{d}#1}
\newcommand{\algName}[0]{\textsc{MPQ}\xspace}
\newcommand{\algNameReal}[1]{\textsc{MPQH#1REAL}\xspace}
\newcommand{\algNameDR}[1]{\textsc{MPQH#1DR}\xspace}

\newcommand{\pendulum}{\textsc{PendulumSwingup}\xspace}
\newcommand{\ballincup}{\textsc{BallInCupSparse}\xspace}
\newcommand{\fetchpush}{\textsc{FetchPushBlock}\xspace}
\newcommand{\frankadrawer}{\textsc{FrankaDrawerOpen}\xspace}

\newcommand{\softQ}{\textsc{SoftQLearning}\xspace}

\renewcommand{\eqref}[1]{Eq.~(\ref{#1})}

\newcommand{\figref}[1]{Fig.~\ref{#1}}
\newcommand{\algoref}[1]{Algorithm~\ref{#1}}

\newtheorem{observation}{O}
\newtheorem{ques}{Q}
 \usepackage[subtle]{savetrees}

\newcommand{\xxnote}[3]{}
\ifx\hidenotes\undefined
\renewcommand{\xxnote}[3]{\color{#2}{#1: #3}}
\fi

\newcommand{\fullFigGap}[0]{\vspace{-1.5\baselineskip}} 

\title{Information Theoretic Model Predictive Q-Learning}
\vspace{-4mm}
\author{
	\centering \Name{Mohak Bhardwaj}\textsuperscript{1,2} \qquad \Name{Ankur Handa}\textsuperscript{2} \qquad \Name{Dieter Fox}\textsuperscript{1,2} \qquad \Name{Byron Boots}\textsuperscript{1,2} \\ 
	\textsuperscript{1}University of Washington \quad \textsuperscript{2}NVIDIA \\
}

\begin{document}

\maketitle
\vspace{-7mm}
\begin{abstract}
Model-free Reinforcement Learning (RL) works well when experience can be collected cheaply and model-based RL is effective when system dynamics can be modeled accurately. However, both assumptions can be violated in real world problems such as robotics, where querying the system can be expensive and real-world dynamics can be difficult to model. 
In contrast to RL, Model Predictive Control (MPC) algorithms use a simulator to optimize a simple policy class online, constructing a closed-loop controller that can effectively contend with real-world dynamics. MPC performance is usually limited by factors such as model bias and the limited horizon of optimization. In this work, we present a novel theoretical connection between information theoretic MPC and entropy regularized RL and develop a Q-learning algorithm that can leverage biased models. We validate the proposed algorithm on sim-to-sim control tasks to demonstrate the improvements over optimal control and reinforcement learning from scratch. Our approach paves the way for deploying reinforcement learning algorithms on real systems in a systematic manner.
\end{abstract}
\vspace{-2mm}
\section{Introduction}\label{sec:introduction}\vspace{-1mm}
Deep reinforcement learning has generated great interest due to its success on a range of difficult problems including Computer Go~\citep{silver2016mastering} and high-dimensional control tasks such as humanoid locomotion~\citep{schulman2015trust,lillicrap2015continuous}. While these methods are extremely general and can learn policies and value functions for complex tasks directly from raw data, they are also sample inefficient, and partially-optimized solutions can be arbitrarily poor, resulting in safety concerns when run on real systems. 

One straightforward way to mitigate these issues is to learn a policy or value function entirely in a high-fidelity simulator~\citep{todorov2012mujoco, airsim2017fsr} and then deploy the optimized policy on the real system. However, this approach can fail due to model bias, external disturbances, the subtle differences between the real robot hardware and poorly modeled phenomena such as friction and contact dynamics. Sim-to-real transfer approaches based on domain randomization (DR)~\citep{tobin2017domain, peng2018sim} and model ensembles~\citep{kurutach2018model,shyam2019model} aim to make the policy robust by training it to be invariant to varying dynamics. However, DR approaches are very sensitive to the choice of distribution, which is often designed by hand. 

Model predictive control (MPC) is a widely used method for generating feedback controllers and has a rich history in robotic control, ranging from aggressive autonomous driving~\citep{williams2017information,wagener2019online} to contact-rich manipulation~\citep{kumar2014real,fu2016one}, and humanoid locomotion~\citep{erez2013integrated}. MPC repeatedly optimizes a finite horizon sequence of controls using an approximate dynamics model that predicts the effect of these controls on the system. The first control in the optimized sequence is executed on the real system and the optimization is performed again from the resulting next state. However, the performance of MPC can suffer due to approximate or simplified models and a limited lookahead. Therefore the parameters of MPC, including the model and horizon $H$ need to be carefully tuned to obtain good performance. While using a longer horizon is generally preferred, real-time requirements may limit the amount of lookahead, and a biased model can result in compounding model errors. In the context of RL, local optimization is an effective way of improving an imperfect value function as noted by~\cite{silver2016mastering,silver2017mastering,sun2018truncated,lowrey2018plan,anthony2017thinking}. However, these approaches assume access to a perfect model, which is not the case when dealing with robotic systems. Hence, we argue that it is essential to learn a value function from real data and utilize local optimization to stabilize learning.

In this work, we present an approach to RL that leverages the complementary properties of model-free reinforcement learning and model-based optimal control. 
Our proposed method views MPC as a way to simultaneously approximate and optimize a local Q function via simulation, and Q-learning as a way to improve MPC using real-world data. 
We focus on the paradigm of entropy regularized reinforcement learning where the aim is to learn a stochastic policy that minimizes the cost-to-go as well as KL divergence with respect to a prior policy. This has been explored in RL and Inverse RL for its better sample efficiency and exploratory properties ~\citep{ziebart2008maximum,todorov2009efficient,haarnoja2017reinforcement,schulman2017equivalence}. This approach also enables faster convergence by mitigating the over-commitment issue in the early stages of Q-learning~\citep{fox2015taming}.  
We discuss how this formulation of reinforcement learning has deep connections to information theoretic stochastic optimal control where the objective is to find control inputs that minimize the cost while staying close to the passive dynamics of the system~\citep{theodorou2012relative}. This helps in both injecting domain knowledge into the controller as well as mitigating issues caused by over optimizing the biased estimate of the current cost due to model error and the limited horizon of 
optimization. We explore this connection in depth and derive an infinite horizon information theoretic model predictive control algorithm based on~\citet{williams2017information}. We test our approach called Model Predictive Q-Learning (\algName) on simulated continuous control tasks and compare it against information theoretic MPC and soft Q-Learning~\citep{haarnoja2017reinforcement}, where we demonstrate faster learning with much fewer system interactions (a few minutes with real system parameters) and better performance compared to MPC and soft Q-Learning even in the presence of sparse rewards. The learned Q-function allows us to truncate the MPC planning horizon, which provides additional computational benefits.
Finally, we also compare \algName against domain randomization on sim-to-sim tasks. We conclude that DR approaches can be sensitive to the hand-designed distributions for randomizing parameters which causes the learned Q-function to be biased and suboptimal on the true system's parameters, whereas learning from data generated on true system is able to overcome biases and adapt to the real dynamics.  

\vspace{-2mm}
\section{Preliminaries}\label{sec:preliminaries}\vspace{-1mm}
We first introduce the entropy-regularized RL and information theoretic MPC frameworks and show that they are complimentary approaches to solve a similar problem. 
\vspace{-3mm}
\subsection{Reinforcement Learning with Entropy Regularization} \label{subsec:entropy_reg_rl}
A Markov Decision Process (MDP) is defined by the tuple $\mathcal{M} = \MDP$ where $\stateSpace$ is the state space, $\actionSpace$ is the action space, $\cost$ is the per-step cost function, $s_{t+1} \sim \prob{ \cdot | s_t, a_t}$ is the stochastic transition dynamics, $\discount$ is the discount factor and $\mu$ is the prior distribution over the initial state. A closed-loop policy $\pi(\cdot|s)$ outputs a distribution over actions given a state. Running $\pi$ on the system for H-steps starting from time $t$ results in a distribution over trajectories denoted by $\trajprob{\pi}{\dynfn}{t}{H}$ with $\tau = (s_t, a_t, \ldots, s_{t+H-1}, a_{t+H-1})$ being a trajectory sample such that $a_{t} \sim \pi(a_t | s_t)$ and $s_{t+1} \sim \prob{s_{t+1} | s_t, a_t}$.
The KL divergence between $\pi$ and a \textit{prior} policy $\prior$ at a particular state is $\KL{\pi(\cdot|s)}{\prior(\cdot|s)} = \expect{\pi}{\log \left(\pi(a|s) / \prior(a|s)\right)}$. Given $c_{t} = c(s_t, a_t)$ and $ \mathrm{KL}_{t} = \KL{\pi(\cdot|s_t)}{\prior(\cdot|s_t)}$, entropy-regularized RL~\citep{fox2015taming} aims to optimize the objective
\vspace{-1mm}
{\small
\begin{equation}
\label{eq:entropyrlobj}
\pi^{*} = \argminprob{\pi}\; \expect{\tau \sim  \trajprob{\pi}{\dynfn}{0}{\infty} }{ \sum_{t=0}^{\infty} \gamma^{t} \left(c_t + \lambda \mathrm{KL}_{t}\right)\; }
\end{equation}
}
where $s_0 \sim \mu$ and $\lambda$ is a temperature parameter that penalizes deviation of $\pi$ from $\prior$. Given $\pi$, we can define the \textit{soft} value functions and their $H$-timestep versions as\footnote{In this work we consider costs instead of rewards and hence aim to find policies that minimize cumulative cost-to-go.}


{\small
\begin{align}
\vspace{-2mm}
\label{eq:softval}
V^{\pi}(s_t)  &= \expect{\tau \sim \trajprob{\pi}{\dynfn}{t}{\infty}}{\sum_{l=0}^{\infty} \gamma^{l}\left( c_{t+l} + \lambda \mathrm{KL}_{t+l} \right)} = \expect{\tau \sim \trajprob{\pi}{\dynfn}{t}{H}}{\sum_{l=0}^{H-1} \gamma^{l}\left(c_{t+l} + \lambda \mathrm{KL}_{t+l}\right) + \gamma^{H} V^{\pi}(s_{t+H})} \\ 
Q^{\pi}(s_t, a_t) &= c_t + \gamma \expect{s_{t+1} \sim \dynfn}{V^{\pi}(s_{t+1})} = c_t + \expect{\tau \sim \trajprob{\pi}{\dynfn}{t+1}{H}}{\sum_{l=1}^{H-1} \gamma^{l} \left(c_{t+l}  + \lambda \mathrm{KL}_{t+l}\right) + \gamma^{H} (\lambda \mathrm{KL}_{t+H} + Q(s_{t+H}, a_{t+H}))} \nonumber
\vspace{-2mm}
\end{align}
}

It can be verified that {\small$V^{\pi}(s_t)  = \expect{a_t \sim \pi_t}{\lambda \log(\pi(a_t|s_t) / \prior(a_t|s_t) + Q(s_t,a_t)}$}. 
The optimization in~\eqref{eq:entropyrlobj} can be performed either by policy gradient methods that aim to find the optimal policy $\pi \in \policyclass$ via stochastic gradient descent~\citep{schulman2017equivalence} or value based methods that try to iteratively approximate the value function of the optimal policy. In either case, the output is a global closed-loop control policy $\pi^{*}(a|s)$.

\vspace{-3mm}
\subsection{Information Theoretic MPC}
Solving the above optimization for a global closed-loop policy can be prohibitively expensive and hard to accomplish during \textit{online} operation, i.e. at every time step, as the system executes, especially when using complex policy classes like deep neural networks. In contrast, MPC computes a closed-loop policy by online optimization of a simple policy class with a truncated horizon. To achieve this, MPC algorithms such as Model Predictive Path Integral Control (MPPI)~\citep{williams2017information} solve a surrogate MDP {\small $\hat{\mathcal{M}} = \MDPSim$} at every timestep with an approximate dynamics model $\dynfncap$, which can be a deterministic simulator such as MuJoCo~\citep{todorov2012mujoco}, and a shorter planning horizon $H$.\footnote{We assume perfect state and cost information, as is common in MPC algorithms~\citep{williams2017information}.} At timestep $t$, starting from the current system state $s_t$, a sequence of actions $A = \left(a_t, a_{t+1}, \ldots a_{t+H-1} \right)$ is sampled from a parameterized open-loop control distribution {\small$\pi_{\theta}(A)$, where $\theta = \left[\theta_t, \theta_{t+1}, \ldots \theta_{t+H-1} \right]^{T}$} is a vector of parameters. Since the actions are independent of state, we consider them to be sampled sequentially $\pi_{\theta}(A) = \pi_{\theta_{t}}(a_t) \prod_{l=t+1}^{t+H-1}\pi_{\theta_{l}}(a_l | a_t, a_{t+1}, \ldots a_{l-1})$. The policy and simulator result in a trajectory distribution {\small $\trajprob{\pi_{\theta}}{\dynfncap}{t}{H}$} with each trajectory {\small $\tau = (s_t, a_t, \ldots, s_{t+H-1}, a_{t+H-1})$} sampled such that at timestep $t+l$, $a_{t+l} \sim \pi_{\theta_{t+l}}(a_{t+l} | a_t, \ldots, a_{t+l-1})$ and $s_{t+l+1} \sim \dynfncap(s_{t+l+1}|s_{t+l}, a_{t+l})$. Algorithms like MPPI aim to find an optimal $\theta^{*}$ that optimizes 
\vspace{-2mm}
{\small
\begin{equation}
\label{eq:klcontrol}
\theta^* = \argminprob{\theta} \; \expect{\tau \sim \trajprob{\pi}{\dynfncap}{t}{H}}{ \sum_{l=0}^{H-2}\gamma^{l} (c_{t+l} +  \lambda \mathrm{KL}_{t+l}) + \gamma^{H-1}\left(c_f(s_{t+H-1}, a_{t+H-1}) + \lambda \mathrm{KL}_{t+H-1} \right)}
\end{equation}
}
where {\small $\mathrm{KL}_{t+l} = \KL{\pi_{\theta}(a_{t+l}|a_t, \ldots, a_{t+l-1})}{\prior_{\phi}(a_{t+l} | a_t, \ldots, a_{t+l-1})}$}, $\prior_{\phi}(A)$ is the passive dynamics of the system, i.e the distribution over actions produced when the control input is zero with parameters $\phi$ and $c_f$ is a terminal cost function. Once $\theta^{*}$ is obtained, the first action from the resulting distribution is executed on the system and the optimization is performed again from the  next state resulting in a closed-loop controller. The re-optimization and entropy regularization helps in mitigating effects of model-bias and inaccuracies in optimization by avoiding overcommitment to the current estimate of the cost. A shortcoming of MPC is the finite horizon which is especially pronounced in tasks with sparse rewards where a short horizon can make the agent extremely myopic. To mitigate this, an approach known as \emph{infinite horizon MPC}~\citep{zhong2013value} sets the terminal cost $c_f$ as a value function that adds global information to the problem.

\vspace{-2mm}
\section{Approach}\label{sec:approach}\vspace{-2mm}
We explore the connection between entropy-regularized RL and MPPI and use it to develop an infinite horizon MPC procedure. This enables us to use MPC to approximate the Q-function and Q-learning from real-data as a way to mitigate finite horizon and model-bias issues inherent with MPC. We first derive the expression for the infinite-horizon optimal policy, which is intractable to sample from and then a scheme to iteratively approximate it with a simple policy class similar to~\citet{williams2017information}. 

\vspace{-2mm}
\subsection{Optimal H-step Boltzmann Distribution}\vspace{-1mm}
Let $\pi(A)$ and $\prior(A)$ be the joint control distribution and prior over $H$-horizon \textit{open-loop} actions respectively, with $\pi_{t} = \pi(a_t)$ and $\pi_{t+l} = \pi(a_{t+l} | a_{t+l-1}, \ldots, a_{t})$. Assuming, $\dynfn$ is deterministic, the following equations hold~\citep{fox2015taming}
\vspace{-5mm}%
{\small
\begin{align}
\label{eq:valfndeterm}
&V^{\pi}(s_t) = \expect{a_t \sim \pi_{t}}{\lambda \log(\pi(a_t) / \prior(a_t) + Q^{\pi}(s_t,a_t)} 
&Q^{\pi}(s_t,a_t) = c_t + \gamma V^{\pi}(s_{t+1}) 
\end{align}
}
We also assume the undiscounted case with discount factor, $\gamma = 1$.~\footnote{Refer to Appendix~\ref{subsec:gammaderiv} for discussion on the discounted case.}
Substituting from the equation for $Q^{\pi}(s,a)$ into $V^{\pi}(s)$ in~\eqref{eq:valfndeterm}
{\small
\vspace{-1mm}
\begin{align*}
V^{\pi}(s_t) &= \expect{a_t \sim \pi_t}{\lambda \log \left(\frac{\pi(a_t)}{\prior(a_t)} \right) + c_t
 + \expect{a_{t+1} \sim \pi_{t+1}}{\lambda \log \frac{\pi(a_{t+1} | a_{t})}{ \prior(a_{t+1} | a_{t})} + Q^{\pi}(s_{t+1}, a_{t+1})}} \\ \nonumber
 &= \expect{a_t \sim \pi_t, a_{t+1} \sim \pi_{t+1}}{c_t + \lambda \left( \log \frac{\pi(a_t)\pi(a_{t+1} | a_{t})}{\prior(a_t)\prior(a_{t+1} | a_{t})} \right) + Q^{\pi}(s_{t+1}, a_{t+1})} \nonumber \\
 &= \mathbb{E}_{a_t, a_{t+1} \sim \pi(a_t, a_{t+1})}\Bigg[c_t + \lambda \left( \log \frac{\pi(a_t, a_{t+1})}{\prior(a_t, a_{t+1})} \right) + Q^{\pi}(s_{t+1}, a_{t+1})\Bigg]
\end{align*}
}
where $c_t$ and $\log \frac{\pi(a_t)}{\prior(a_t)}$ are taken inside the expectation as they are constants with respect to $\pi(a_{t+1} \mid a_t)$. Recursing $H$ times, 
{\small
\vspace{-1mm}
\begin{align}
\label{eq:valopenloop}
V^{\pi}(s_t) &= \expect{(a_t \ldots a_{t+H-1}) \sim \pi(A)}{\sum_{l=0}^{H-2}c_{t+l}+ \lambda \sum_{l=0}^{H-1} \log \frac{\pi(a_{t+l} | a_{t+l-1} \ldots a_t)}{\prior(a_{t+l} | a_{t+l-1} \ldots a_t)} + Q^{\pi}(s_{t+H-1}, a_{t+H-1})} \nonumber \\
&= \expect{(a_t \ldots a_{t+H-1}) \sim \pi(A)}{\sum_{l=0}^{H-2} c_{t+l} + \lambda \log \frac{\pi(A)}{\prior(A)} + Q^{\pi}(s_{t+H-1}, a_{t+H-1})}
\vspace{-2mm}
\end{align}
}
~\eqref{eq:valopenloop} is similar to~\eqref{eq:softval} with the key difference being the use of open-loop policies and the deterministic dynamics assumption, leading to the expectation and KL divergence being applied to the joint action distribution rather than the state-action trajectory distribution $\trajprob{\pi}{\dynfn}{t}{H}$. Now, consider the following joint action distribution over horizon $H$\vspace{-2mm}
{\small
\begin{equation}
\label{eq:policyopt}
\pi(A) = \frac{1}{\eta} \exp{\left(\frac{-1}{\lambda}\left(\sum_{l=0}^{H-2}c_{t+l} + Q^{\pi}(s_{t+H-1}, a_{t+H-1})\right)\right)}\overbar{\pi}(A)
\vspace{-2mm}
\end{equation}
}
where {\small $\eta = \expect{\overbar{\pi}(A)}{\exp{\left(\frac{-1}{\lambda}\left(\sum_{l=0}^{H-2}c_{t+l} + Q^{\pi}(s_{t+H-1}, a_{t+H-1})\right)\right)}}$} is a normalizing constant.
We show that this is the optimal control distribution as $\nabla V^{\pi}(s_t) = 0$. Substituting ~\eqref{eq:policyopt} into~\eqref{eq:valopenloop})
{\small
\begin{equation*}
\vspace{-2mm}
V^{\pi}(s_t) = \expect{\pi(A)}{\sum_{l=0}^{H-2} c_t -\lambda \log(\eta) - \sum_{l=0}^{H-2}c_t - Q^{\pi}(s_{t+H-1}, a_{t+H-1}) + Q^{\pi}(s_{t+H-1}, a_{t+H-1})}  
= \expect{\pi(A)}{- \lambda \log(\eta)}
\end{equation*}
}
Since $\eta$ is a constant,  $V^{\pi}(s) = -\lambda \log(\eta)$. Therefore, for $\pi$ in~\eqref{eq:policyopt}, the soft value function is a constant with gradient zero and is thus the optimal value function, i.e
{\small
\begin{equation}
\label{eq:hstepfreeenergy}
V^{\pi^{*}}(s_t) = - \lambda \log \expect{\overbar{\pi}(A)}{\mathrm{exp}\left(\frac{-1}{\lambda}\left(\sum_{l=0}^{H-2}c(s_{t+l}, a_{t+l}) + Q^{\pi}(s_{t+H-1}, a_{t+H-1})\right)\right)}
\end{equation}
}
which is often referred to in optimal control literature as the ``free energy" of the system~\citep{theodorou2012relative}. For H=1,~\eqref{eq:hstepfreeenergy} takes the form of the soft value function from~\citet{haarnoja2018soft}.

\vspace{-2mm}
\subsection{Infinite Horizon MPPI Update Rule}\label{subsec:infmppi}\vspace{-1mm}
Here we derive our infinite horizon MPPI update rule following the approach of ~\cite{williams2017information}. Since sampling actions from the optimal control distribution in~\eqref{eq:policyopt} is intractable, we consider parameterized control policies $\pi_{\theta}(A) \in \policyclass$ which are easy to sample from. We then optimize for a vector of $H$ parameters $\theta$, such that the resulting action distribution minimizes the KL divergence with the optimal policy\vspace{-1mm}
{\small
\begin{equation}	
	\vspace{-2mm}
	\label{eq:mppiopt}
	\theta^{*} = \argminprob{\theta}{\; \KL{\policyopt(A)}{\pi_{\theta}(A)}}
\end{equation}
}
The objective can be expanded as \vspace{-1mm}
{\small
\vspace{-1mm}
\begin{equation}
\KL{\policyopt(A)}{\pi_{\theta}(A)} \;=\; \indefint{A}\; \policyopt(A) \log \frac{\policyopt(A)}{\pi_{\theta}(A)} \D{A}\;
\;=\; \indefint{A}\; \policyopt(A) \left(\log \frac{\policyopt(A)}{\prior(A)} -  \log \frac{\pi_{\theta}(A)}{\prior(A)}\right)\;\D{A}
\end{equation}
\vspace{-1mm}
\begin{equation}
\label{eq:mppioptfinal}
\theta^{*} = \argmaxprob{\pi_{\theta}(A)}{\;\indefint{A}\; \policyopt(A) \log \frac{\pi_{\theta}(A)}{\prior(A)}\; \D{A}}
\end{equation}
}
where first term was removed as it was independent of $\theta$. Consider $\policyclass$, to be a time-independent multivariate Gaussian over sequence of the $H$ controls with constant covariance $\Sigma$ at each timestep. We can write control distribution and prior as\vspace{-2mm}
{\small
\begin{align}
\label{eq:gaussianpolicies}
\pi_{U}(A) & = \frac{1}{Z}\prod_{t=0}^{H-1}\exponent{-\frac{1}{2}\left(u_t - a_t \right)^{T}\Sigma^{-1}\left(u_t - a_t \right)}
&\prior(A) = \frac{1}{Z}\prod_{t=0}^{H-1} \exponent{-\frac{1}{2}a_{t}^{T}\Sigma^{-1}a_{t}}
\end{align}}
where $u_t$ and $a_t$ are the control inputs and actions respectively at timestep $t$ and $Z$ is the normalizing constant. Here the prior corresponds to the passive dynamics of the system~\citep{theodorou2012relative, williams2017information}, although other choices are possible. The policy parameters $\theta$ are the sequence of control inputs $U = \left[u_1, u_2, \ldots, u_H \right]$, which is the mean of the Gaussian. Substituting in~\eqref{eq:mppioptfinal}:\vspace{-2mm}
{\small
\begin{equation}
\label{eq:mppioptgaussian}
U^{*} = \argmaxprob{\pi_{U}(A)}{\;\int\; \policyopt(A)\left(\sum_{t=0}^{H-1}-\frac{1}{2}u_{t}^{T}\Sigma^{-1}u_t + u_{t}^{T}\Sigma^{-1}a_{t} \right)\; \D{A}}
\end{equation}}
The objective can be simplified to the following by integrating out the probability in the first term\vspace{-2mm}
{\small
\begin{equation}
\sum_{t=0}^{H-1} -\frac{1}{2}u_{t}^{T}\Sigma^{-1}u_{t} + u_{t}^{T}\int\policyopt(A) \Sigma^{-1}a_{t} \; \D{A} 
\end{equation}
}
Since this is a concave function with respect to every $u_t$, we can find the maximum by setting its gradient with respect to $u_t$ to zero and solving for the optimal $u_{t}^{*}$\vspace{-2mm}
{\small
\begin{align}
\label{eq:optcontrol}
u_{t}^{*} &= \int \policyopt(A)a_t\D{A}
          &= \int \pi_{U}(A)\frac{\policyopt(A)}{\prior(A)}\frac{\prior(A)}{\pi_{U}(A)}a_{t}\D{A} 
          &= \expect{\pi_{U}(A)}{\frac{\policyopt(A)}{\prior(A)}\frac{\prior(A)}{\pi_{U}(A)}a_{t} } = \expect{\pi_{U}(A)}{w(A)a_{t} } 
\end{align}
}
where the second equality comes from importance sampling to convert the optimal controls into an expectation over the control distribution instead of the optimal distribution, which is intractable to sample from. The importance weight $w(A)$ can be written as follows (substituting $\policyopt$ from ~\eqref{eq:policyopt})\vspace{-2mm}
{\small
\begin{equation}
\label{eq:impweight}
w(A) =  \frac{1}{\eta}\;\exp{\left(\frac{-1}{\lambda}\left(\sum_{t=0}^{H-2}c(s_t, a_t) + Q^{\policyopt}(s_{H-1}, a_{H-1})\right)\right)\frac{\prior(A)}{\pi_{U}(A)}}
\end{equation}
}

~\eqref{eq:optcontrol} gives the expression for the mean of the optimal distribution (or optimal control inputs) as the expectation over the control distribution $\pi_{U}$ of the weighted actions, with weights given by~\eqref{eq:impweight}. In practice, we estimate this expectation using a finite number of Monte-Carlo samples from current control distribution $\pi_{U}$ and iteratively update its mean towards the optimal. To make this clearer we make a change of variables $u_t + \epsilon_t = a_t$ for noise sequence $\mathcal{E} = \left(\epsilon_0 \ldots \epsilon_{H-1} \right)$ sampled from independant Gaussians with zero mean and covariance $\Sigma$ similar to~\citet{williams2017information} and get


{\footnotesize
\begin{align}
\label{eq:imweightnoise}
w(\mathcal{E}) &= \frac{1}{\eta} \;\exp{\left(\frac{-1}{\lambda}\left(\sum_{t=0}^{H-2}c(s_t, u_t + \epsilon_t) + \lambda \log \left(\frac{\pi(U+\mathcal{E})}{\prior(U+\mathcal{E})}\right) + Q^{\policyopt}(s_{H-1}, u_{H-1} + \epsilon_{H-1})\right)\right)} \nonumber \\
& = \frac{1}{\eta} \;\mathrm{exp}\Bigg(\frac{-1}{\lambda}\left(\sum_{t=0}^{H-2}c(s_t, u_t + \epsilon_t) + \lambda \sum_{t=0}^{H-1}\frac{1}{2}u_{t}^{T}\Sigma^{-1}(u_{t} + 2\epsilon_t) +  Q^{\policyopt}(s_{H-1}, u_{H-1} + \epsilon_{H-1})\right)\Bigg)
\end{align}
}
where $\eta$ can be estimated from $N$ Monte-Carlo samples as
{\footnotesize
\begin{equation}
\label{eq:etamc}
	\eta =  \sum_{n=1}^{N}\mathrm{exp}\Bigg(\frac{-1}{\lambda}\Bigg(\sum_{t=0}^{H-2}c(s_t, u_t + \epsilon_t^{n}) + \lambda \sum_{t=0}^{H-1}\frac{1}{2}u_{t}^{T}\Sigma^{-1}(u_{t} + 2\epsilon_t^{n}) +  Q^{\policyopt}(s_{H-1}, u_{H-1} + \epsilon_{H-1}^{n})\Bigg)\Bigg)
\end{equation}
}
We can now form the following iterative update rule where at every iteration $i$ the sampled control sequence is updated according to 
\begin{equation}
	\label{eq:infhorizonmppiupdate}
	u_{t}^{i+1} = u_{t}^{i} + \alpha \sum_{n=1}^{N}w(\mathcal{E}_n)\epsilon_t^n
\end{equation}
where $\alpha$ is a step-size parameter proposed by~\citet{wagener2019online}. 
~\eqref{eq:infhorizonmppiupdate} is the infinite horizon MPPI update rule. For $H=1$, it corresponds to soft Q-learning with stochastic optimization to find the optimal action. This leads us to our soft Q-learning algorithm, \algName that uses infinite horizon MPPI to generate actions and Q-targets.

\vspace{-2mm}
\subsection{Information Theoretic Model Predictive Q-Learning}
We consider parameterized value functions $Q_{\theta}(s,a)$ where parameters $\theta$ are updated by stochastic gradient descent on the loss {\small $L(\theta) = \frac{1}{K}\sum_{i=1}^{K}(y_i - Q_{\theta}(s_i,a_i))^{2}$} for a batch of $K$ experience tuples $(s, a, c, s')$ sampled from a replay buffer. Targets $y_{i}$ are calculated using the Bellman equation as 
{\footnotesize
\begin{equation}
\label{eq:value_target}
y = c(s,a)  - \gamma\lambda \log \mathrm{E}_{\policyopt(A)} \Bigg[\mathrm{exp}\Bigg(\frac{-1}{\lambda}\Bigg(\sum_{t=0}^{H-2}c(s_t, a_t) + \sum_{t=0}^{H-1} \lambda \log \frac{\pi^*_{t}}{\prior_t} + Q_{\theta}(s_{H-1}, a_{H-1})\Bigg)\Bigg)\Bigg|s_0=s'\Bigg ] 
\end{equation}
}%
The second term is the same as free energy from~\eqref{eq:hstepfreeenergy} with the expectation over prior converted to expectation over the optimal policy $\pi^{*}$ using importance sampling.
Since the value function updates are performed offline, we can utilize large amounts of computation to obtain $\pi^{*}(A)$. We do so by performing multiple iterations of the infinite horizon MPPI update in~\eqref{eq:infhorizonmppiupdate} from $s'$,
which allows for better approximation of the free energy (akin to approaches such as Covariance Matrix Adaption, although MPPI does not adapt the covariance). This  helps in early stages of learning by providing better quality targets than a random Q function. Intuitively, this update rule leverages the biased dynamics model $\dynfncap$ for $H$ steps and a soft Q function at the end learned from interactions with the real system. 

At every timestep $t$ during online rollouts, an $H$-horizon sequence of actions is optimized using infinite horizon MPPI and the first action is executed on the system. Online optimization with predictive models can look ahead to produce better actions than ad-hoc exploration strategies such as $\epsilon$-greedy. Combined with the better Q estimates, this helps in accelerating learning as we demonstrate in our experiments.
~\algoref{alg:mpq} shows the complete \algName algorithm.
A closely related approach in literature is POLO~\citep{lowrey2018plan}, which also uses MPPI and offline value function learning, however they do not explore the connection between MPPI and entropy regularized RL, and thus the algorithm does not use free energy targets.

\vspace{-1mm}
\begin{algorithm}
	\caption{\algName \label{alg:mpq}}
	\SetKwInput{Input}{Input}
	\SetKwInput{Parameter}{Parameter}
	\Input{Approximate model $\dynfncap$, initial Q function parameters $\theta_1$,  experience buffer $\dataset$}
	\Parameter{Number of episodes $N$, length of episode $T$, planning horizon $H$, number of update episodes $N_{update}$, minibatch-size $K$, number of minibatches $M$}
	\For{$i=1 \ldots N$}{
		\For{$t=1 \ldots T$}{
			$(a_t,\ldots,a_{t+H}) \leftarrow$ Infinite horizon MPPI (\eqref{eq:infhorizonmppiupdate}) \\ \label{line:policy}
			 Execute $a_t$ on the real system to obtain $c(s_{t}, a_{t})$ and next state $s_{t+1}$ \\
			$\dataset \leftarrow (s_t, a_t, c, s_{t+1}) $\\
		    } 
	    \If{$i \% N_{update} == 0$}{
	    	Sample $M$ minibatches of size $K$  from $\dataset$ \\
    		Generate targets using~\eqref{eq:value_target} and update parameters to $\theta_{i+1}$\\
    	}
       \KwRet $\theta_N$ or best $\theta$ on validation.
		}
\end{algorithm}

\vspace{-4mm}
\section{Experiments}\label{sec:experiments}
\vspace{-2mm}
We evaluate the efficacy of \algName on two fronts: (a) overcoming the shortcomings of both stochastic optimal control and model free RL in terms of computational requirements, model bias, and sample efficiency; and (b) learning effective policies on systems for which accurate models are not known. 
\vspace{-2mm}
\subsection{Experimental Setup}\label{subsec:exp_setup} 
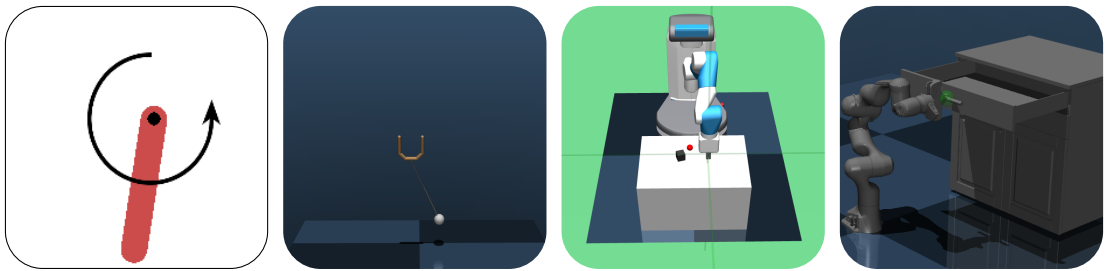
\begin{figure}
\centering
\begin{tikzpicture} 
\begin{scope}
    \clip [rounded corners=.5cm] (0,0) rectangle coordinate (centerpoint) (3.5cm,3.5cm); 
    \node [inner sep=0pt] at (centerpoint) {\includegraphics[trim=9 0 39 60, clip, width=7.0cm]{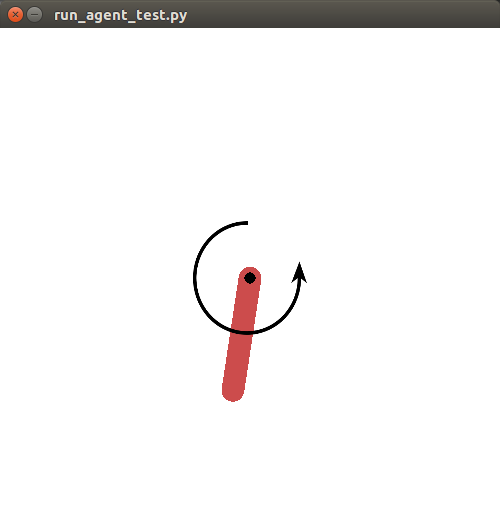}};
    \draw [black, thick, rounded corners=.5cm] (0,0) rectangle coordinate (centerpoint) (3.5cm,3.5cm);
\end{scope}

\begin{scope}[xshift=3.7cm]
    \clip [rounded corners=.5cm] (0,0) rectangle coordinate (centerpoint) ++(3.5cm,3.5cm); 
    \node [inner sep=0pt] at (centerpoint) {\includegraphics[trim=9 0 39 100, clip, width=7.0cm]{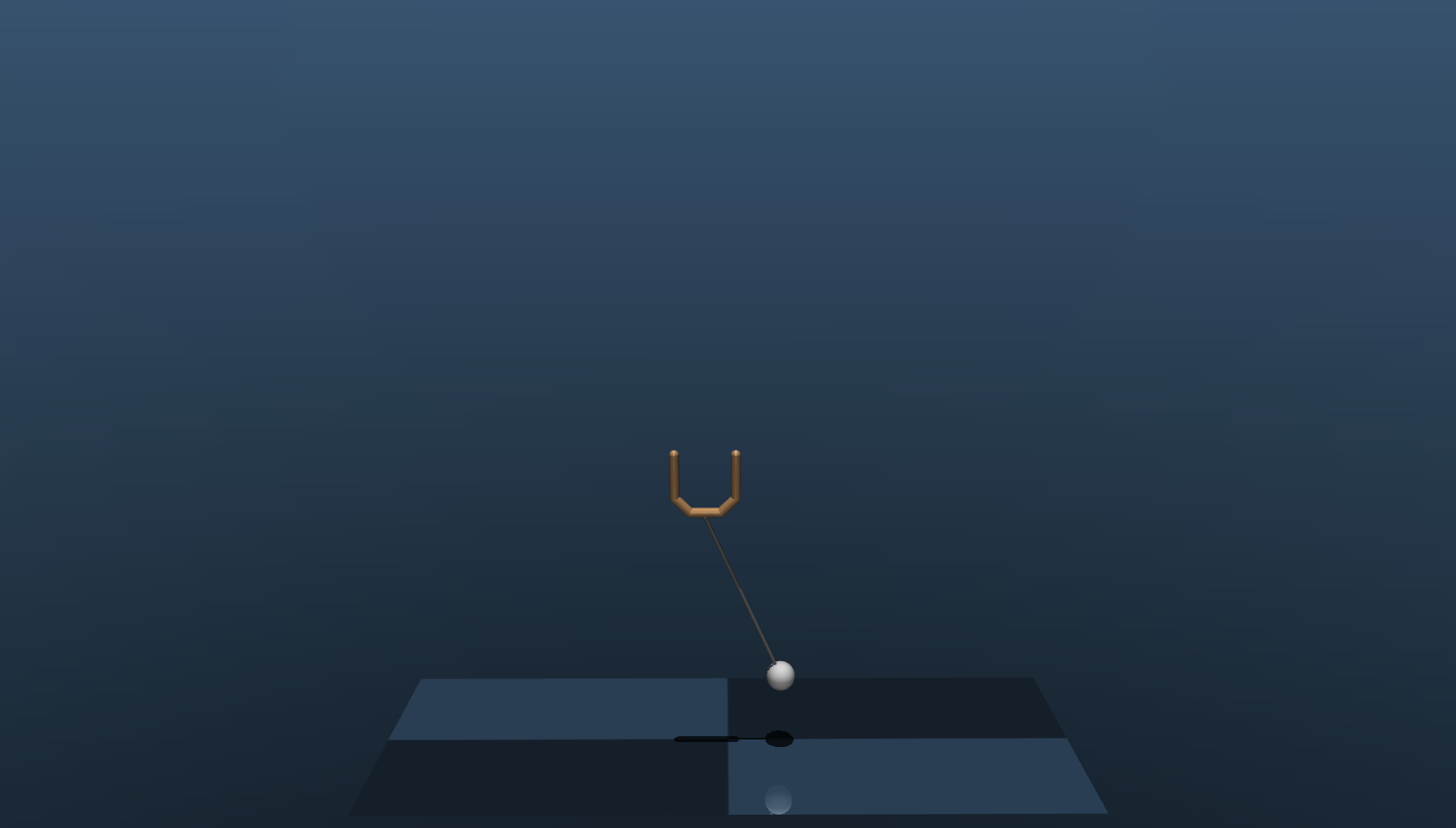}}; 
\end{scope}

\begin{scope}[xshift=7.4cm]
    \clip [rounded corners=.5cm] (0,0) rectangle coordinate (centerpoint) ++(3.5cm,3.5cm); 
    \node [inner sep=0pt] at (centerpoint) {\includegraphics[trim=10 150 10 30, scale=0.15]{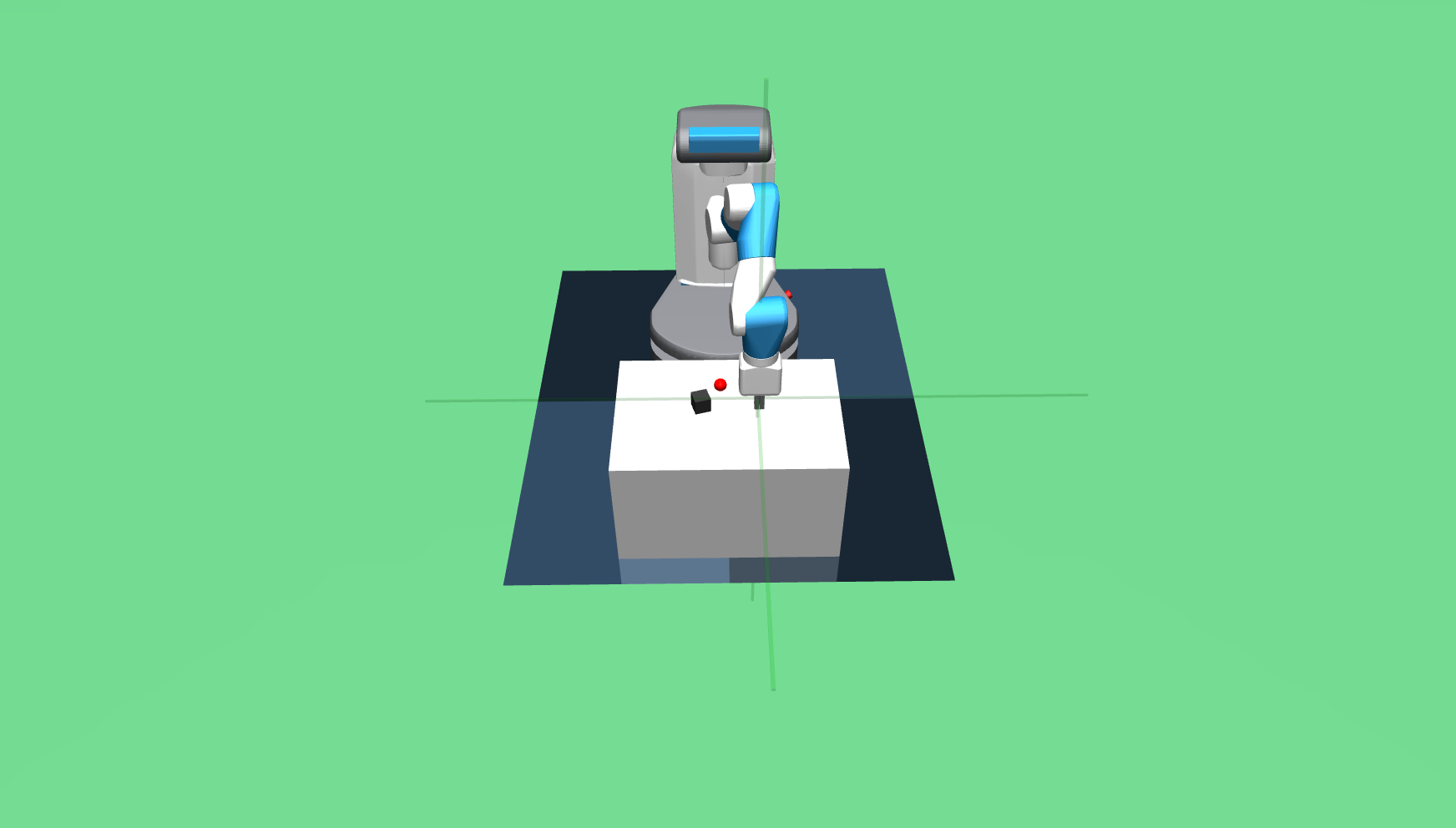}};
\end{scope}

\begin{scope}[xshift=11.1cm]
    \clip [rounded corners=.5cm] (0,0) rectangle coordinate (centerpoint) ++(3.5cm,3.5cm); 
    \node [inner sep=0pt] at (centerpoint) {\includegraphics[trim=0 0 30 0,scale=0.4]{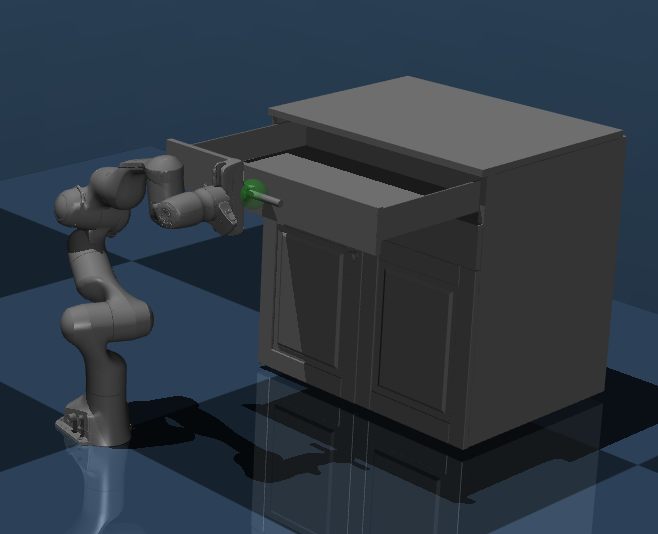}};
\end{scope}

\end{tikzpicture}
\caption{Tasks considered for analyzing the performance of \algName. Refer to Section~\ref{subsec:exp_setup} for details. We consider continous control tasks which require long-horizon planning for success. The agents are provided with a biased distribution over dynamics parameters such as inertia, friction coefficients and joint dampings as a reasonable approximation of model-bias since, these parameters are hard to estimate accurately.}
\label{fig:env_snaps}
\end{figure}

\vspace{-1mm}
We focus on sim-to-sim continuous control tasks using MuJoCo (except \pendulum that uses dynamics equations) to study the properties of \algName in a controlled manner. We consider robotics-inspired tasks (shown in~\figref{fig:env_snaps}) with either sparse rewards or requiring long-horizon planning. The complexity is further aggravated as the agent is not provided with the true dynamics parameters, but a uniform distribution over them with a biased mean and added noise. 
Details of the tasks are 
\begin{enumerate}[wide, labelwidth=!, labelindent=0pt]
	\item \pendulum: the agent tries to swingup and stabilize a pendulum by applying torque on the hinge given a biased distribution over mass and length. Initial state is randomized after every 10s episode. \vspace{-2mm}
	\item \ballincup: a sparse version of the task from the Deepmind Control Suite~\citep{deepmindcontrolsuite2018}. Given a cup and ball attached by a tendon, the goal is to swing and catch the ball. The agent is provided with a biased distribution over the ball's mass, moment of inertia and tendon stiffness. A cost of 1 is incurred at every timestep and 0 if the ball is in the cup which corresponds to success. The position of the ball is randomized after every episode of 4s duration.\vspace{-2mm}
	\item \fetchpush: proposed by~\cite{1802.09464}, the agent controls the cartesian position and opening of a Fetch robot gripper to push a block to a goal location. A biased distribution over the mass, moment of inertia, friction coefficients and size of the object is provided. An episode is successful if the agent gets the block within 5cm of the goal in 4s. The positions of both block and goal are randomized after every episode. \vspace{-2mm}
	\item \frankadrawer: based on a real-world manipulation problem from~\cite{chebotar2019closing} where the agent velocity controls a 7DOF Franka Panda arm to open a cabinet drawer. A biased distribution over damping and frictionloss of robot and drawer joints is provided. Every episode lasts 4s after which the arm configuration is randomized. Success is opening the drawer within 1cm of a target displacement.\vspace{-2mm}
\end{enumerate}

\vspace{-1mm}

The tasks we chose are more realistic proxies for real-world robotics tasks than standard OpenAI Gym~\citep{1606.01540} baselines such as \textsc{Ant} and \textsc{HalfCheetah}. The parameters we randomize are reasonable in real world scenarios as estimating moment of inertia and friction coefficients is especially error prone. Details of default parameters and randomization distributions are provided in~\tableref{tab:environment_details}. Experiments were performed on a desktop with 12 Intel Core i7-3930K @ 3.20GHz CPUs and 32 GB RAM with only few hours of training. Q-functions are parameterized with feed-forward neural networks that take as input an observation vector and action. Refer to Appendix~\ref{subsec:moreexperiments} for further details.

\begin{figure}[!t]
	\centering
	\subfigure[\pendulum]{
		\centering
		\includegraphics[trim=9 4 39 34, clip, width=0.4\textwidth]{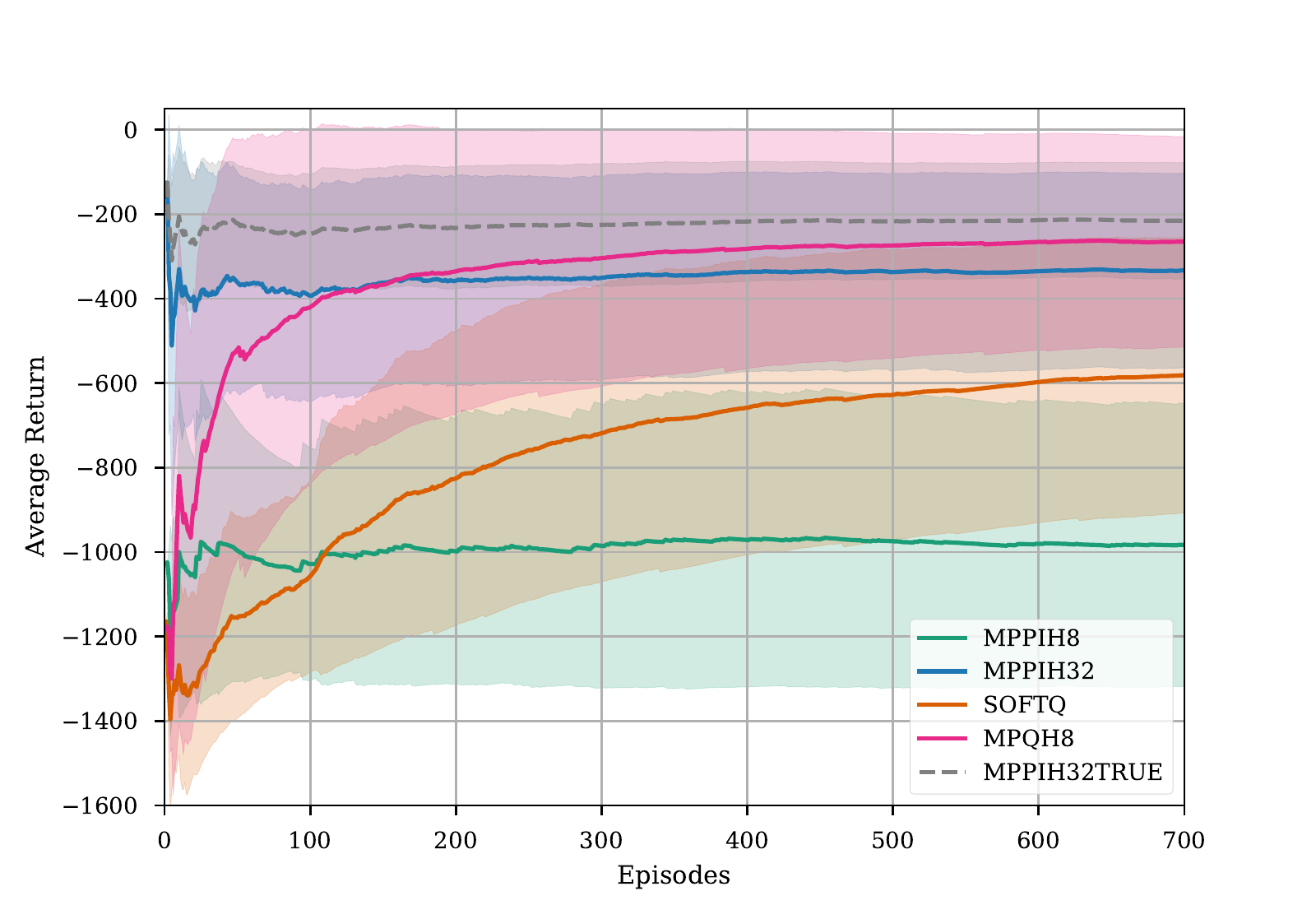}}%
	\subfigure[\ballincup]{
		\centering
		\includegraphics[trim=9 4 39 34, clip, width=0.4\textwidth]{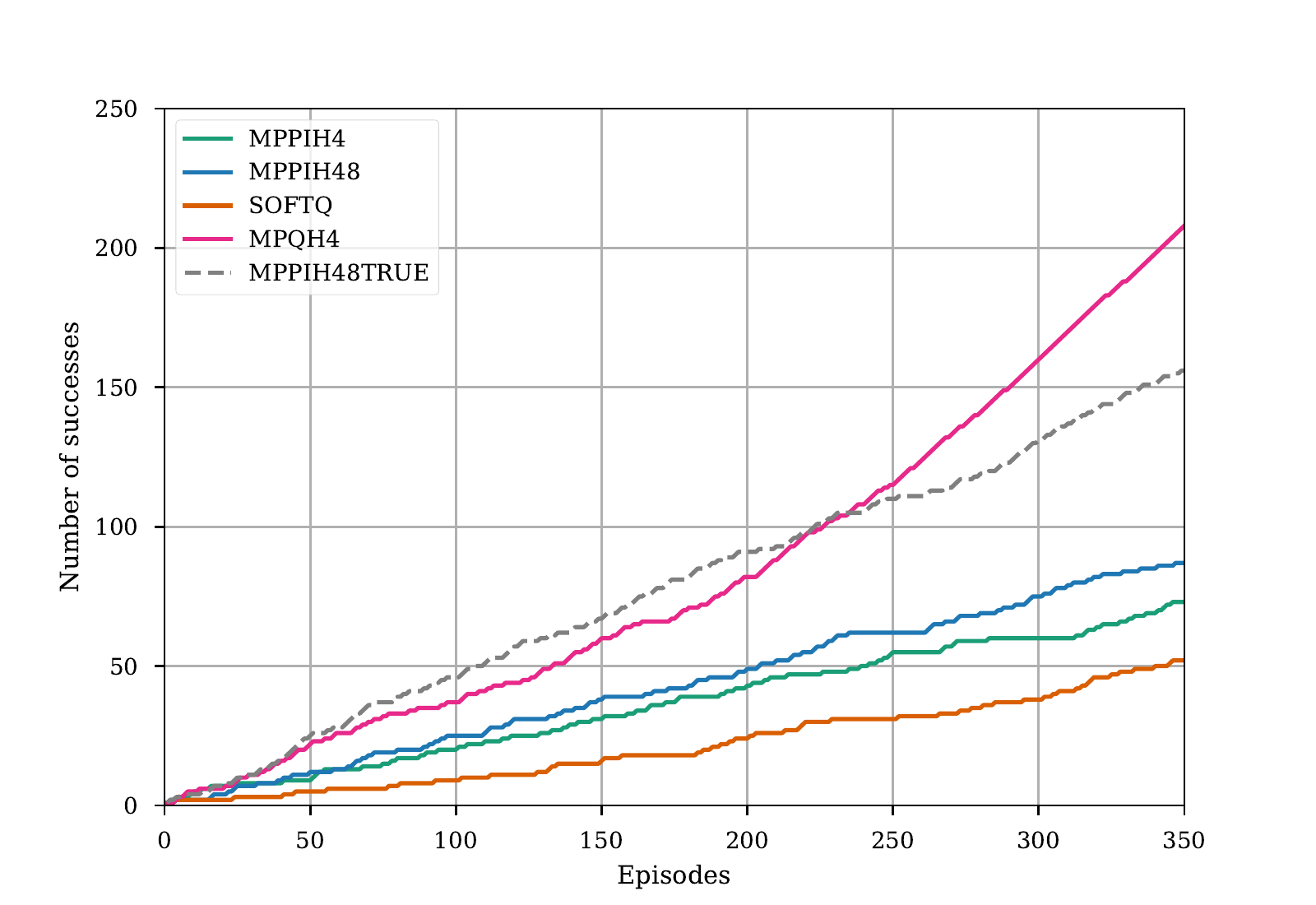}}%

	\centering
	\subfigure[\fetchpush]{
		\centering
		\includegraphics[trim=9 4 39 34, clip, width=0.4\textwidth]{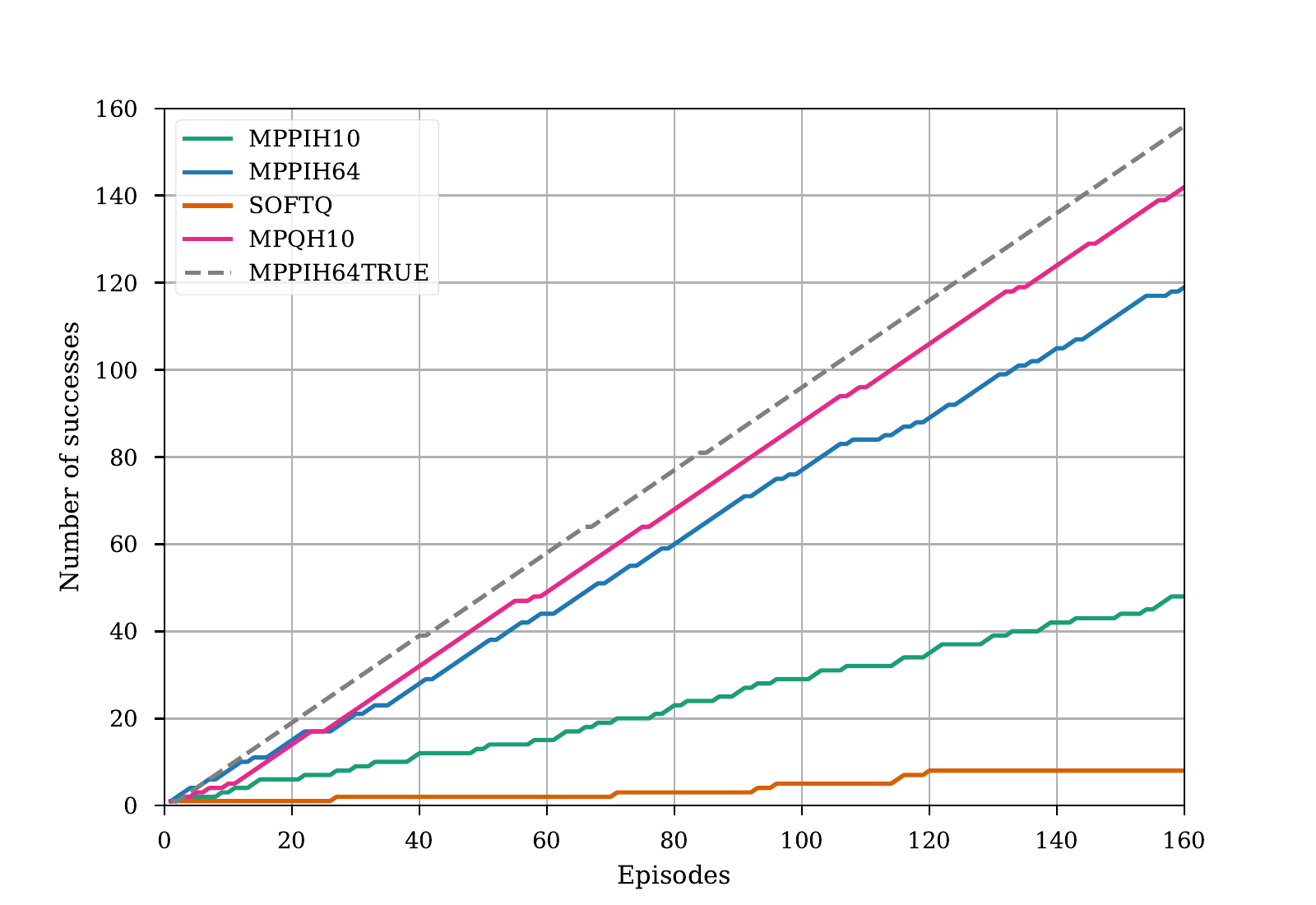}}%
	\subfigure[\frankadrawer]{
		\centering
		\includegraphics[trim=9 4 39 34, clip, width=0.4\textwidth]{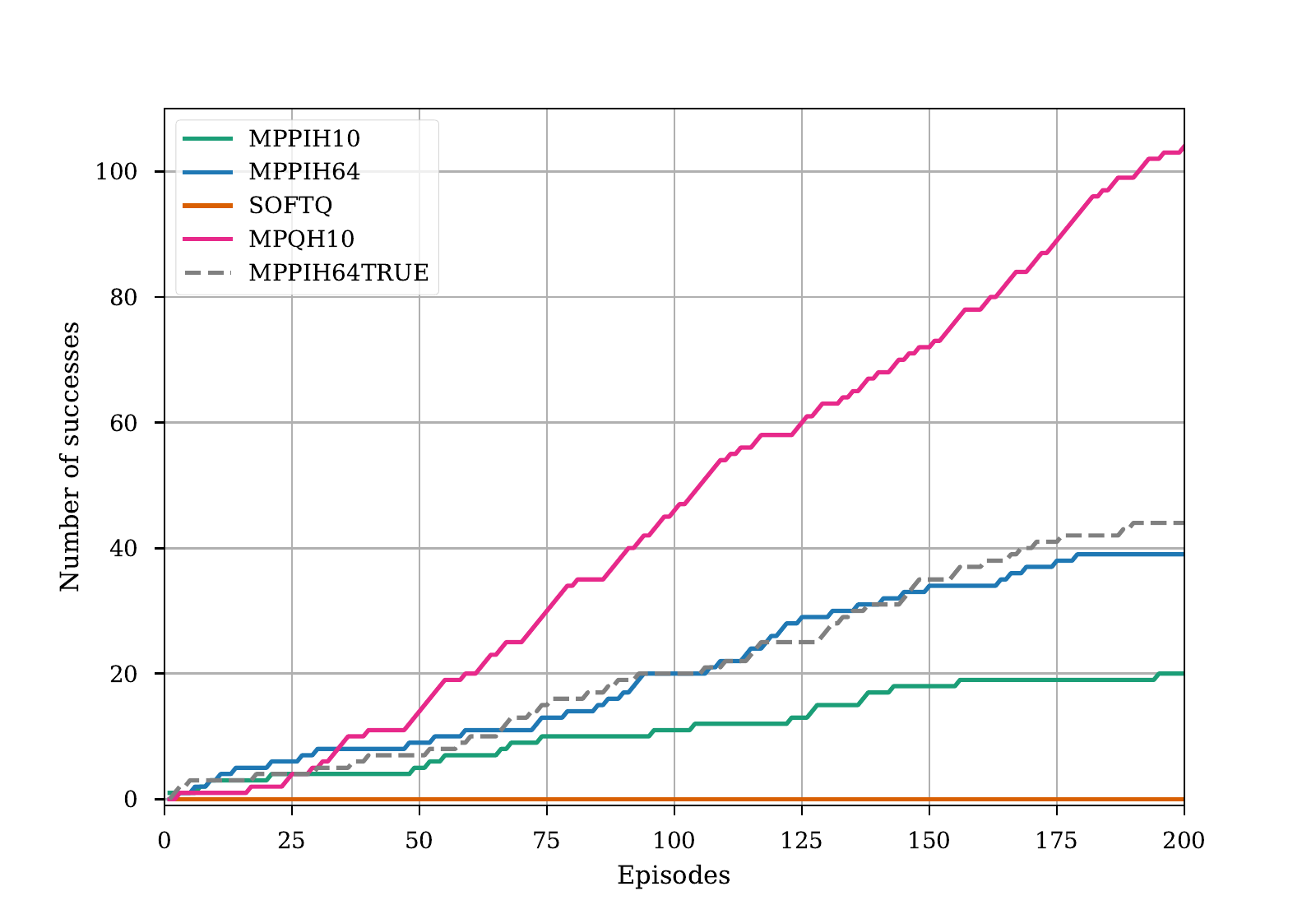}}%
\vspace{-2mm}
	\caption{Comparison of \algName against baselines during training. The number following $H$ in the legend corresponds to MPC horizon. The dashed grey line shows performance of MPPI with access to true dynamics and no terminal Q function, denoting the upper limit on the performance MPPI can achieve. The horizon for MPPI and \algName were selected after a rough grid search.}


	\label{fig:trainingcomparison}
	\vspace{-4mm}
\end{figure}

\begin{table}[!hbtp]
	\centering
	\caption{Environment parameters and dynamics randomization. The last column denotes the range for the uniform distribution. $I_{xyz}$ implies that moment of inertia is the same along all three axes. $T$ is the tendon stiffness. For \fetchpush, the block is a cube with sides of length $l$. \fetchpush and \frankadrawer use uniform distribution for every parameter given by: $\text{mean} = \text{bias}\times\text{true value}$ and $\text{range} = [-\sigma, \; \sigma] \times \text{true\_value}$  }
	\label{tab:environment_details}
	\begin{tabular}{|c|c|c|c|}
		\hline
		\textbf{Environment} & \textbf{Cost Function} & \textbf{True Parameters} & \textbf{Biased Distribution}\\ \hline
		
		\multirow{2}{*}{\pendulum} & \small $\Theta^{2} + 0.1\dot{\Theta}^{2}$ & \small $ m = 1\text{kg}$ & \small $ m=[0.9, 1.5]$ \\ &&\small $l = 1\text{m}$ &  \small $l=[0.9, 1.5]$ \\ \hline
		
		\multirow{3}{*}{\ballincup} & \multirow{3}{*}{\small $\text{0 if ball in cup}$} &\small $m = 0.058 \text{kg}$ & \small $m=[0.0087, 0.87]$ \\ & & \small $I_{xyz}=1.47\times10^{-5}$ & \small $I_{xyz}=[0.22,22]\times10^{-5}$ \\  &\small 1 else & \small $T=0.05$ & \small  $T=[0.00375, 1.5]$\\ \hline
		
		\multirow{4}{*}{\fetchpush} & &\small $m = 2 \text{kg}$ & \small $\text{bias}=0.35$ \\ &  \small $d_{block, goal}$ &\small $I_{xyz}= 8.33e-4$ & \small $\sigma = 0.45$ \\ & & \small $\mu=[1, 0.005, 10^{-4}]$ & \small \\ & & \small $l=0.025m$ & \\ \hline

		\multirow{2}{*}{\frankadrawer}& \tiny $d_{ee,h} + 0.08d^{ang}_{ee,h}$ &\small frictionloss = 0.1 & \small $\text{bias}=0.1$ \\ & \tiny $-1.0 + d_{drawer}/d_{max}$ & \small damping=0.1 & \small $\sigma = 10.0$ \\\hline
	\end{tabular}
\end{table}

\vspace{-2mm}
\subsection{Analysis of Overall Performance}\vspace{-1mm}
By learning a terminal value function from real data we posit that \algName will adapt to true system dynamics and allow us to truncate the MPC horizon. Using MPC for Q targets, we also expect to require significantly less data than model-free Q-learning. Hence, we compare \algName with the following natural baselines: vanilla MPPI with same horizon as \algName, MPPI with longer horizon, MPPI with longer horizon + true dynamics and \softQ with target networks. 
Note \algName does not use a target network. We do not compare with model-based RL methods~\citep{kurutach2018model,chua2018deep} as learning globally consistent neural network models adds an additional layer of complexity beyond the scope of this work. MPQ is complementary to model learning and one can benefit from the other. We make following observations: 

\begin{observation}
\vspace{-3mm}
\algName can truncate the planning horizon leading to computational efficiency over MPPI.
\vspace{-1mm}
\end{observation}
~\figref{fig:trainingcomparison} shows that \algName outperforms MPPI with the same horizon after only a few training episodes and ultimately outperforms MPPI with a much longer horizon. This is due to global information encapsulated in the Q-function, hardness of optimizing longer sequences and compounding model error in longer rollouts. In \fetchpush, MPPI with a short horizon ($H=10$) is unable to reach near the block whereas \algName with $H=10$ outperforms MPPI with $H=64$ within 30 training episodes i.e. about 2 minutes of interaction with true sim parameters. 
In the high-dimensional \frankadrawer, \algName with $H=10$ achieves a success rate of $>$5X MPPI with $H=10$, and outperforms MPPI with $H=64$ within a few minutes of interaction.
We also examine the effects of varying the horizon during training and present the result of an ablation study in Appendix~\ref{subsec:horizon_comparison}.
\begin{observation}
\vspace{-1mm}
\algName mitigates effects of model-bias through a combination of MPC, entropy regularization and a Q function learned from true system.
\vspace{-1mm}
\end{observation}
~\figref{fig:trainingcomparison} shows that \algName with short horizon achieves performance close to, or better than, MPPI with true dynamics and a longer horizon (dashed gray line) in all tasks.
\begin{observation}
\vspace{-1mm}
Using MPC provides stable Q targets leading to sample efficiency over \softQ
\vspace{-1mm}
\end{observation}
In \ballincup, \fetchpush and \frankadrawer, \softQ does not converge to a consistent policy whereas \algName achieves good performance within few minutes of interaction with true system parameters.

\vspace{-2mm}
\subsubsection*{Case Study: Learning Policies for Systems With Inaccurate Models}
DR aims to make a policy learned in simulation robust by randomizing the simulation parameters. But, such policies can be suboptimal with respect to true parameters due to bias in randomization distribution. 
\vspace{-2mm}
\begin{ques}
Can a Q-function learned using rollouts on a real system overcome model bias and outperform DR?
\end{ques}
\vspace{-1mm}
We compare against a DR approach inspired by~\citet{peng2018sim} where simulated rollouts are generated by sampling different parameters at every timestep from a broad distribution shown in~\tableref{tab:environment_details} whereas real system rollouts use the true parameters.
Table~2 shows that a Q function learned using DR with only simulated experience is unable to generalize to the true parameters during testing and \algName has over 2X the success rate in \ballincup and 3X in \frankadrawer. 
Note that MPC always uses simulated rollouts, the difference is whether the data for learning the Q function is generated using biased simulation (in DR approach) or true parameters. 
\begin{wraptable}{r}{0.53\linewidth}
	\vspace{-13mm}
 	\centering
 	\caption*{\scriptsize {\bf Table 2}: Average success when training Q function using real system rollouts (ending with REAL) and DR. Test episodes=100. $H$ is horizon of MPC in both training and testing, with $H=1$ being \softQ}
 	\label{tab:sim_to_real_results}
 	\vspace{-2mm}
 	\begin{tabular}{|c|c|c|}
 		\hline
 		\scriptsize \textbf{Task} & \scriptsize  \textbf{Agent} & \scriptsize \textbf{Avg. success rate} \\
 		\hline
 		\multirow{4}{*}{\scriptsize  \ballincup}& \scriptsize \algNameReal{4} & \scriptsize  \cmark{0.85} \\
 		&\scriptsize \algNameDR{4} &\scriptsize  0.41  \\
		&\scriptsize \algNameReal{1}&\scriptsize  0.09 \\
 		\scriptsize (350 training episodes)&\scriptsize \algNameDR{1} &\scriptsize  0.06 \\
		\hline
		\multirow{4}{*}{\scriptsize \frankadrawer}&\scriptsize \algNameReal{10} & \scriptsize \cmark{0.53} \\
 		 &\scriptsize \algNameDR{10} & \scriptsize 0.17  \\
		&\scriptsize \algNameReal{1}& \scriptsize 0.0 \\
 		\scriptsize (200 training episodes)&\scriptsize \algNameDR{1} & \scriptsize 0.0 \\
		\hline
 	\end{tabular}
 \end{wraptable}



\vspace{-3mm}
\section{Discussion}\label{sec:discussion}
\vspace{-1mm}
We presented a theoretical connection between information theoretic MPC and entropy-regularized RL that naturally provides an algorithm to leverage the benefits of both. While the approach is effective on a range of tasks, in the future we wish to investigate the dependence between model error and MPC horizon and adapt the horizon by reasoning about the quality of the Q function, both critical for real-world applications.
\acks{The authors would like to thank Nolan Wagener for insightful discussions for improving the manuscript.}

\bibliography{references.bib}
\newpage
\section*{Appendix} \label{sec:appendix}

\renewcommand{\thesubsection}{\Alph{subsection}}
\subsection{Optimal $H$-Step Boltzmann Distribution with Discounting}\label{subsec:gammaderiv}

As before, let $\pi(A)$ and $\prior(A)$ be the joint control distribution and prior over $H$-horizon \textit{open-loop} actions respectively, with $\pi_{t} = \pi(a_t)$ and $\pi_{t+l} = \pi(a_{t+l} | a_{t+l-1}, \ldots, a_{t})$. Continuing from~\eqref{eq:valopenloop} with discount factor $\gamma$ we have
{\small
\begin{equation}
\label{eq:valopenloopdisc}
V^{\pi}(s_t) = \expect{(a_t \ldots a_{t+H-1}) \sim \pi(A)}{\sum_{l=0}^{H-2} \gamma^{l}c_{t+l}+ \lambda\sum_{l=0}^{H-1} \gamma^{l} \mathrm{KL}_{t+l} + \gamma^{H-1} Q^{\pi}(s_{t+H-1}, a_{t+H-1}))}
\end{equation}
}
where we replaced $\expect{\pi_{t+l}}{\log \frac{\pi(a_{t+l} | a_{t+l-1} \ldots a_t)}{\prior(a_{t+l} | a_{t+l-1} \ldots a_t)}}$ by 
$\mathrm{KL}_{t+l}$. The above value function discounts the future KL divergence terms with $\gamma^{l}$. Now consider the case where we do not discount the $\mathrm{KL}$ divergence terms. Then the long-term value of a policy is given by - 
{\small
\begin{equation}
\label{eq:valopenloopdisc2}
V'^{\pi}(s_t) = \expect{(a_t \ldots a_{t+H-1}) \sim \pi(A)}{\sum_{l=0}^{H-2} \gamma^{l}c_{t+l}+ \lambda\sum_{l=0}^{H-1}\mathrm{KL}_{t+l} + \gamma^{H-1} Q^{\pi}(s_{t+H-1}, a_{t+H-1}))}
\end{equation}
}
Since $\gamma \leq 1$, we can see that $V^{\pi}(s_t) \leq V'^{\pi}(s_t)$, i.e $V'^{\pi}(s_t)$ is an upper-bound on $V^{\pi}(s_t)$. Now, we have
{\small
\begin{equation}
\label{eq:valopenloopdisc2joint}
V'^{\pi}(s_t) = \expect{(a_t \ldots a_{t+H-1}) \sim \pi(A)}{\sum_{l=0}^{H-2} \gamma^{l}c_{t+l}+ \lambda \log \frac{\pi(A)}{\prior(A)} + \gamma^{H-1} Q^{\pi}(s_{t+H-1}, a_{t+H-1}))}
\end{equation}
}
We will find an optimal policy for $V'^{\pi}(s_t)$ which is an upper-bound on the original value function. Consider a H-horizon discounted joint distribution given by
{\small
\begin{equation}
\label{eq:policyoptdisc}
\pi(A) = \frac{1}{\eta} \exp{\left(\frac{-1}{\lambda}\left(\sum_{l=0}^{H-2}\gamma^{l}c_{t+l} + \gamma^{H-1} Q^{\pi}(s_{t+H-1}, a_{t+H-1})\right)\right)}\overbar{\pi}(A)
\vspace{-2mm}
\end{equation}
}
where
\begin{equation}
\eta = \expect{\overbar{\pi}(A)}{\exp{\left(\frac{-1}{\lambda}\left(\sum_{l=0}^{H-2}\gamma^{l}c_{t+l} + \gamma^{H-1}Q^{\pi}(s_{t+H-1}, a_{t+H-1})\right)\right)}}
\end{equation}
Substituting from~\eqref{eq:policyoptdisc} into~\eqref{eq:valopenloopdisc2joint} 
{\small
\begin{align}
\vspace{-2mm}
V'^{\pi}(s_t) &= \expect{\pi(A)}{\sum_{l=0}^{H-2} \gamma^l c_t -\lambda \log(\eta) - \sum_{l=0}^{H-2}\gamma^l c_t - \gamma^{H-1} Q^{\pi}(s_{t+H-1}, a_{t+H-1}) + \gamma^{H-1} Q^{\pi}(s_{t+H-1}, a_{t+H-1})}  \\\nonumber
&= \expect{\pi(A)}{- \lambda \log(\eta)}
\end{align}
}
which is a constant. Hence the policy is~\eqref{eq:policyoptdisc} is the optimal policy with respect to the upper bound $V'^{\pi}$ to the true discounted value function $V^{\pi}$. The rest of the derivation for MPPI update rule follows as usual. The insight here is that in the discounted case, using MPPI actually optimizes an upper-bound to the original entropy-regularized cost function.

\vspace{-2mm}

\subsection{Further experimental details}\label{subsec:moreexperiments}

The Q function takes as input the current action and an observation vector defined as: 
{\small
\begin{enumerate}[wide, labelindent=0pt, labelindent=0pt]
	\item {\small \pendulum}: $\left[\mathtt{cos}(\Theta), \mathtt{sin}(\Theta), \dot{\Theta}\right]$ (3 dim)
	
	\item {\small \ballincup}: $\left[x_{\texttt{ball}}, x_{\texttt{target}}, \dot{x}_{\texttt{ball}}, \dot{x}_{\texttt{target}}, x_{\texttt{target}}-x_{\texttt{ball}}, \mathtt{cos}(\Theta), \mathtt{sin}(\Theta)\right]$ (12 dim) where $\Theta$ is angle of line joining ball and target.
	
	\item {\small \fetchpush}: [$x_{\texttt{gripper}}, \; x_{\texttt{obj}}, \; x_{\texttt{obj}}-x_{\texttt{grip}}, \texttt{gripper\_opening}, \texttt{rot}_{\texttt{obj}}, \dot{x}_{\texttt{obj}}, \omega_{\texttt{obj}}, \\ \texttt{gripper\_opening\_vel}, \dot{x}_{gripper}, \texttt{d(gripper, obj)}, \texttt{x}_{\texttt{goal}}-\texttt{x}_{\texttt{obj}}, \texttt{d(goal,obj)}, x_{\texttt{goal}} $] (33 dim)
	
	\item {\small \frankadrawer:[$ x_{\texttt{ee}}, x_{\texttt{h}}, x_{\texttt{h}}-x_{\texttt{ee}}, \dot{x}_{\texttt{ee}}, \dot{x}_{\texttt{h}}, \texttt{quat}_{\texttt{ee}}, \texttt{quat}_{\texttt{h}}, \texttt{drawer}_{\texttt{disp}},  \texttt{d(ee,h)}, \texttt{d}^\texttt{{quat}(ee,h)}, \texttt{d}^{\texttt{ang}}_{\texttt{ee,h}} $] (39 dim)}
\end{enumerate}
}
For all our experiments we parameterize Q functions with feedforward neural networks with two layers containing 100 units each and $\mathtt{tanh}$ activation. We use Adam~\citep{kingma2014adam} optimization with a learning rate of 0.001. For generating value function targets in~\eqref{eq:value_target}, we use 3 MPPI iterations except 1 for \frankadrawer. The MPPI parameters used are listed in Table~3. 
\vspace{-1mm}
\begin{figure}[t]
	\setcounter{subfigure}{0}
	\subfigure[  \small \ballincup \label{a}]{ 
		\centering
		\includegraphics[trim=9 4 39 34, clip, width=0.33\textwidth]{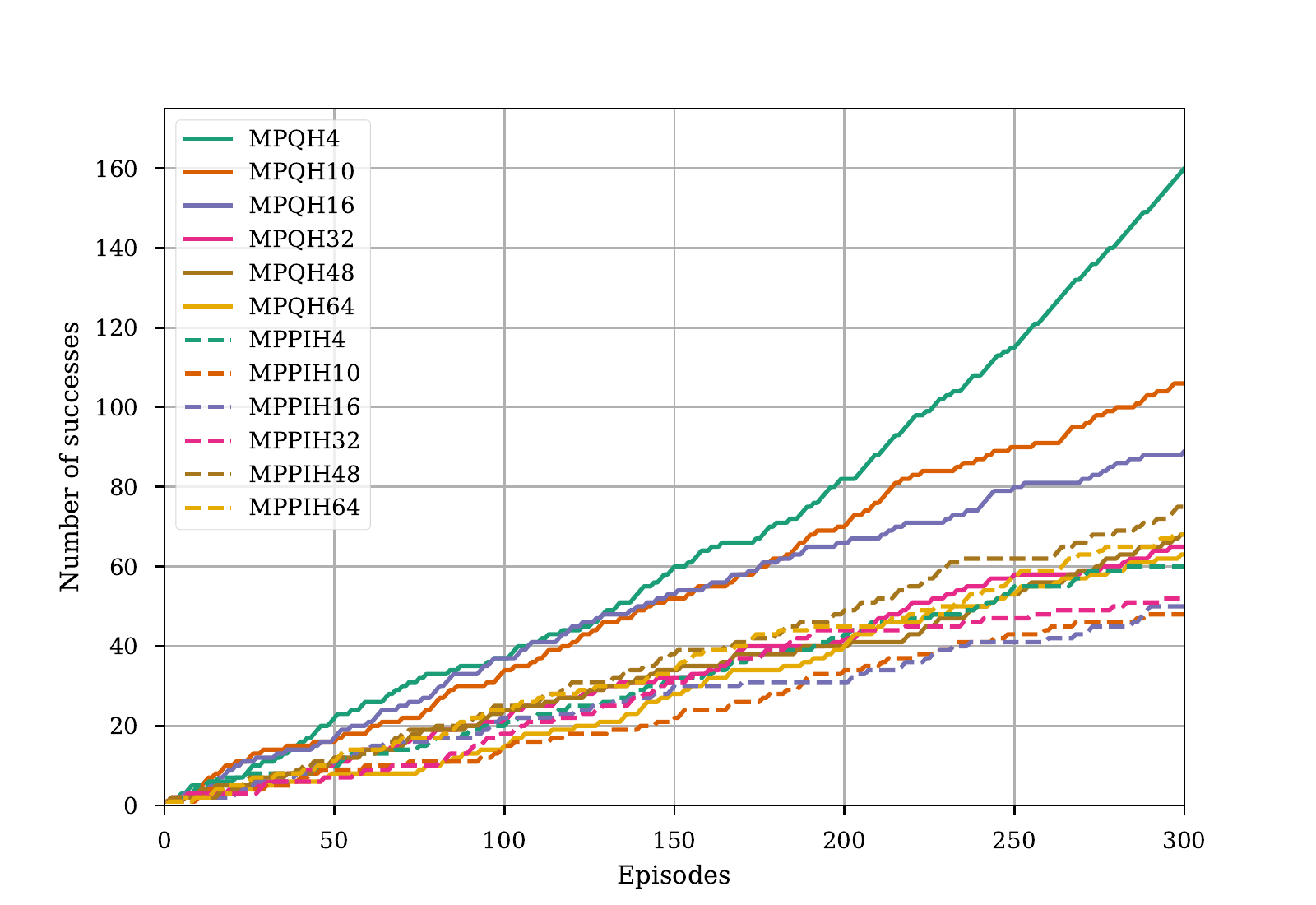}}%
	\subfigure[ \small \fetchpush \label{b}]{ 
		\centering
		\includegraphics[trim=9 4 39 34, clip, width=0.33\textwidth]{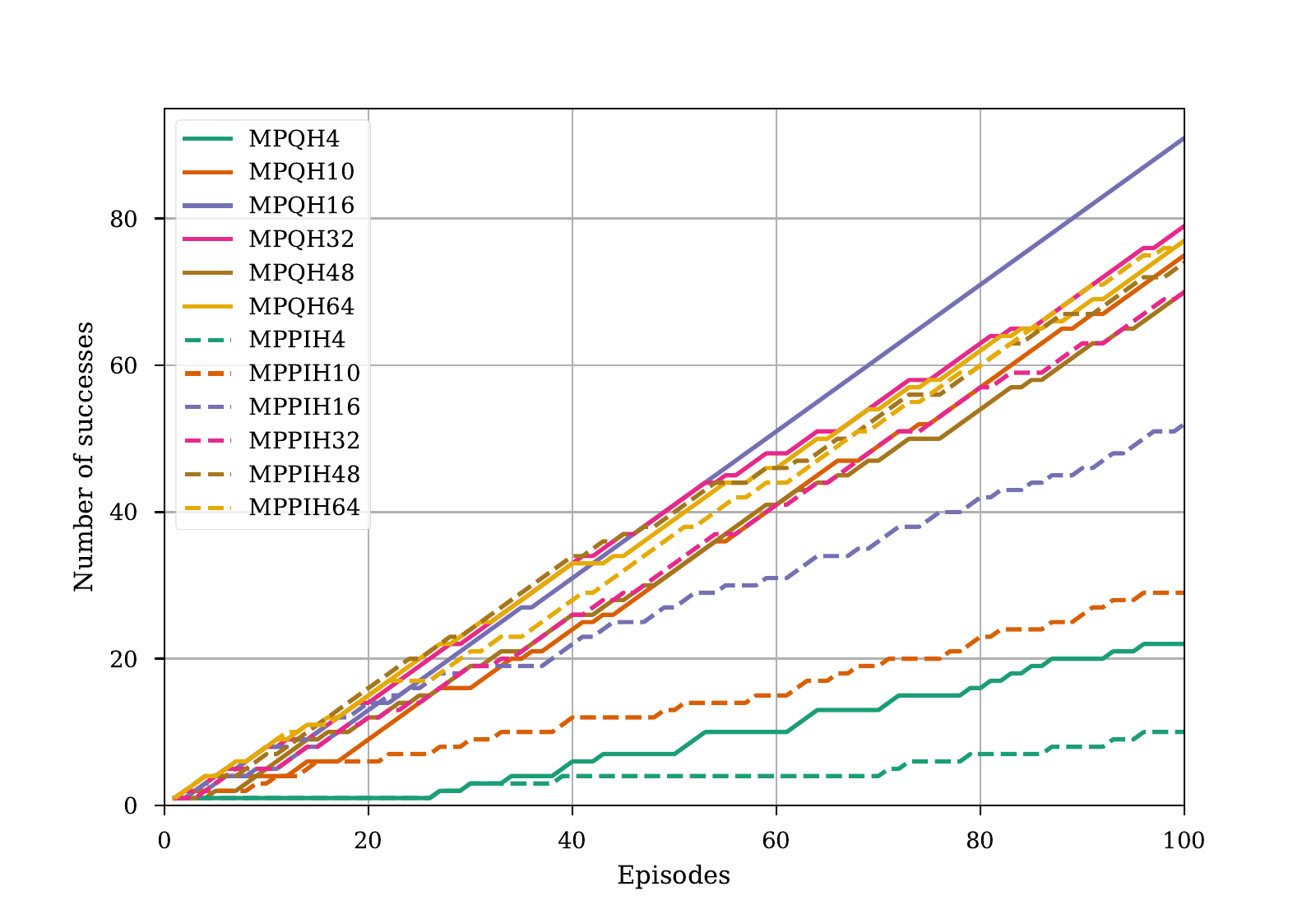}}%
	\subfigure[ \small \frankadrawer \label{c}]{ 
		\centering
		\includegraphics[trim=9 4 39 34, clip, width=0.33\textwidth]{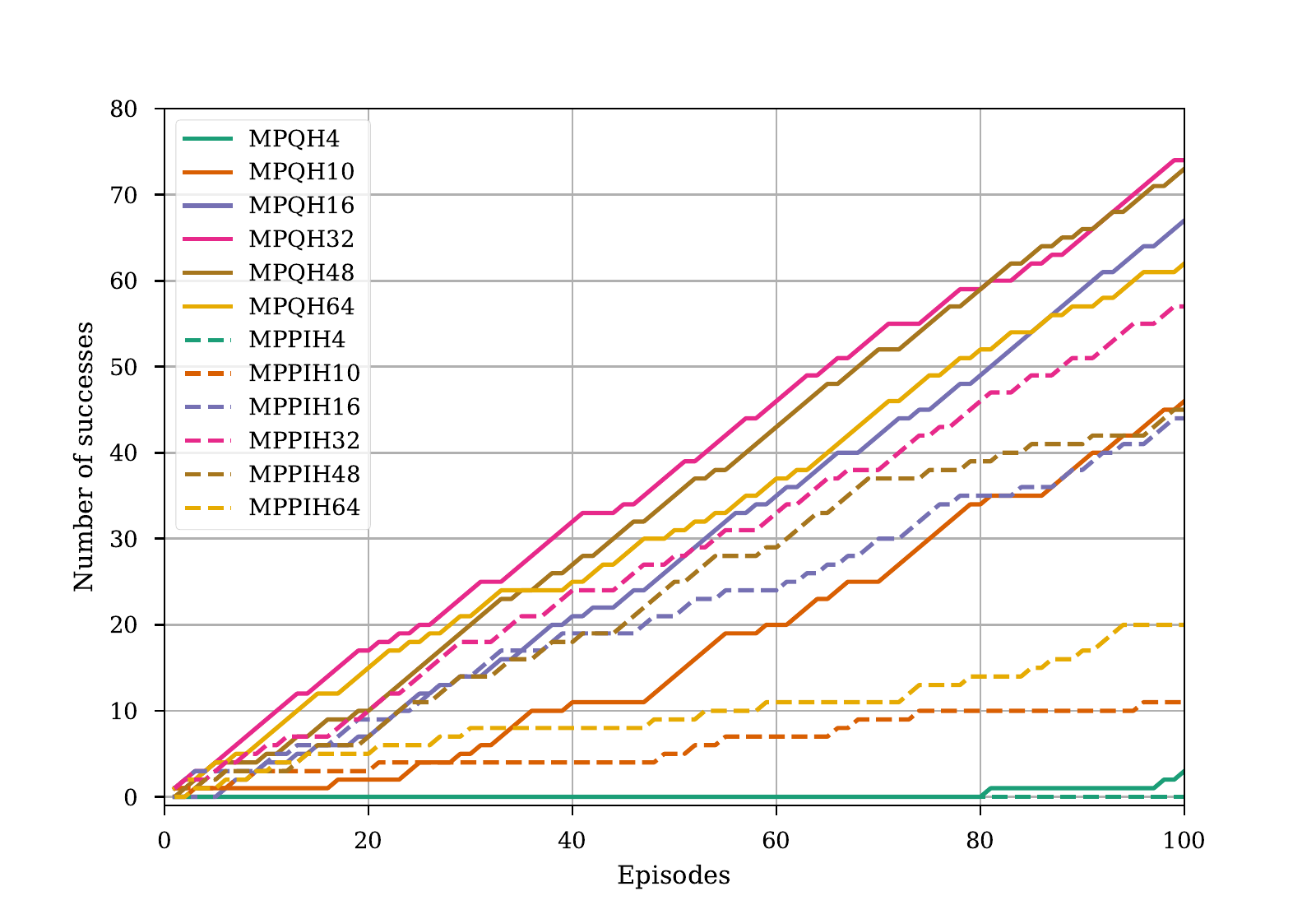}}%
	\vspace{-2mm}
	\caption{\small Effect of varying MPC horizon. The number following $H$ in the legend corresponds to MPC horizon. \fullFigGap}
	\label{fig:horizon_plot}
\end{figure}
\vspace{-2mm}

\vspace{-1mm}
\begin{table}[!hbtp]
	\centering
	\caption*{\small {\bf Table 3:} Cost function and MPPI parameters}
	\vspace{-1mm}
	\centering
	\label{tab:mppi_params}
	\begin{tabular}{|c|c|c|c|c|c|c|}
		\hline
		\textbf{Environment} & \textbf{Cost function} & \textbf{Samples} & $\Sigma$ & $\lambda$ & $\alpha$ & $\gamma$ \\ \hline
		
		\pendulum & \small $\Theta^{2} + 0.1\dot{\Theta}^{2}$ & \small 24 & \small 4.0 & \small 0.15 & \small 0.5 & \small 0.9 \\ \hline
		
		\multirow{3}{*}{\ballincup} & \multirow{3}{*}{\small $\text{0 if ball in cup}$} & \multirow{3}{*}{\small 36}  & \multirow{3}{*}{4.0} & \small \multirow{3}{*}{0.15} & \small \multirow{3}{*}{0.55} & \small \multirow{3}{*}{0.9} \\ &&&&&& \\ & \small 1 else & \small & \small &&& \\ \hline

		\fetchpush & \small $d_{block, goal}$ & \small 36  & 3.0 & {0.01} & \small 0.5 & \small 0.9 \\  \hline

		\multirow{3}{*}{\frankadrawer} & \multirow{3}{*}{\tiny $d_{ee,h} + 0.08d^{ang}_{ee,h}$} & \multirow{3}{*}{\small 36}  & \multirow{3}{*}{4.0} & \small \multirow{3}{*}{0.05} & \small \multirow{3}{*}{0.55} & \small \multirow{3}{*}{0.9} \\ &&&&&& \\ & \tiny $-1.0 + d_{drawer}/d_{max}$& \small & \small &&& \\ \hline
	\end{tabular}	
\end{table}

\vspace{-3mm}
\subsection{Effect of MPC Horizon on Performance}\label{subsec:horizon_comparison}

The planning horizon is critical to the performance of MPC algorithms. We test the effect of the horizon on the performance by running \algName training for different values of $H$ and comparing against MPPI without a terminal value function. The results in~\figref{fig:horizon_plot} provide the following key takeaway
\vspace{-2mm}
\begin{observation}
Longer planning horizon can improve performance but also suffers greatly from model bias and optimization issues
\vspace{-2mm}
\end{observation}

In \ballincup where a very broad range of dynamics parameters is used (see~\tableref{tab:environment_details}) \algName with $H=4$ has the best performance which subsequently degrades with increasing $H$. \fetchpush and \frankadrawer show a trend where performance initially improves with increasing horizon but starts to degrade after a certain point due to compounding effects of model bias. This phenomenon indicates that finding the optimal sweet spot for the horizon is an interesting research direction which we wish to explore thoroughly in the future.

\end{document}